\documentclass[10pt,twocolumn,letterpaper]{article}

\usepackage{cvpr}
\usepackage{times}
\usepackage{epsfig}
\usepackage{graphicx}
\usepackage{amsmath}
\usepackage{amssymb}
\usepackage{multirow}
\usepackage{bm}
\usepackage{pifont}
\usepackage{hhline}
\usepackage{color}
\usepackage{soul}

\usepackage[pagebackref=true,breaklinks=true,letterpaper=true,colorlinks,bookmarks=false]{hyperref}

\cvprfinalcopy 

\def\cvprPaperID{609} 

\ifcvprfinal\fi

\newcommand{\cmark}{\ding{51}}%
\newcommand{\xmark}{\ding{55}}%
\begin{document}

\title{Efficient Video Object Segmentation via Network Modulation}

\author{Linjie Yang$^1$~~~~~~~~~Yanran Wang$^2$\thanks{This work is done when Yanran was an intern at Snap Inc.}~~~~~~~~~Xuehan Xiong$^3$~~~~~~~~~Jianchao Yang$^1$~~~~~~~~~Aggelos K. Katsaggelos$^2$\\
$^1$Snap Inc.\\
$^2$Image and Video Processing Laboratory, Northwestern University\\
$^3$Google Inc.\\
{\tt\small \{linjie.yang,jianchao.yang\}@snap.com \{joycewang1026@u,aggk@eecs\}.northwestern.edu xiong828@gmail.com}
}

\maketitle

\begin{abstract}
Video object segmentation targets at segmenting a specific object throughout a video sequence, given only an annotated first frame. Recent deep learning based approaches find it effective by fine-tuning a general-purpose segmentation model on the annotated frame using hundreds of iterations of gradient descent. Despite the high accuracy these methods achieve, the fine-tuning process is inefficient and fail to meet the requirements of real world applications. We propose a novel approach that uses a single forward pass to adapt the segmentation model to the appearance of a specific object. Specifically, a second meta neural network named modulator is learned to manipulate the intermediate layers of the segmentation network given limited visual and spatial information of the target object. The experiments show that our approach is $70\times$ faster than fine-tuning approaches while achieving similar accuracy. Our model and code are released at \url{https://github.com/linjieyangsc/video_seg}.\vspace{-2pt}
\end{abstract}

\section{Introduction}
\begin{figure}[t]\centering
\includegraphics[width=1.0\linewidth]{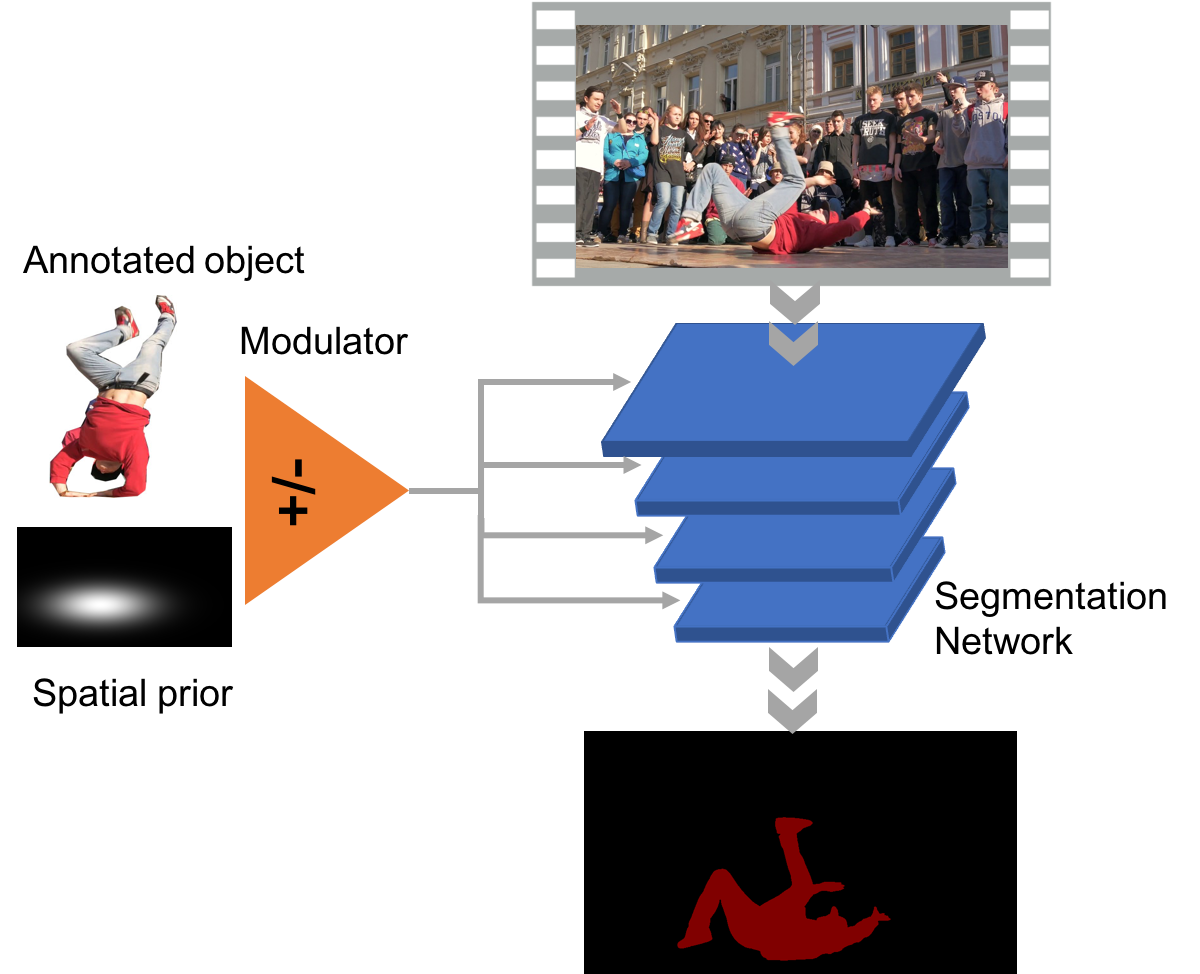}
\caption{An overview of our approach. Our model is consisted of a modulator and a segmentation network. The modulator can adapt the segmentation model instantly to segment an arbitrary object through a video sequence.}
\label{fig:overview}\vspace{-2pt}
\end{figure}

Semantic segmentation plays an important role in understanding visual content of an image as it assigns pre-defined object or scene labels to each pixel and thus translates the image into a segmentation map.
When dealing with video content, a human can easily segment an object in the whole video without knowing its semantic meaning, which inspired a research topic named semi-supervised video segmentation. In a typical scenario of semi-supervised video segmentation, one is given the first frame of a video along with an annotated object mask, and the task is to accurately locate the object in all following frames~\cite{Perazzi2016davis,Li2013video}.
The ability of performing accurate pixel-level video segmentation with minimum supervision (e.g., one annotated frame) can foster a large amount of applications, such as accurate object tracking for video understanding, interactive video editing, augmented reality, and video-based advertisement. When the supervision is limited to only one annotated frame, researchers refer to this scenario as \textit{one-shot learning}. In the recent years, we have witnessed a rising amount of interests in developing one-shot learning techniques for video segmentation~\cite{Caelles2017osvos, Perazzi2017masktrack, Tsai2016objflow, Marki2016bilateral, Shin2017pixel, Cheng2017segflow}. Most of these work share a similar two-stage paradigm: first, train a general-purpose Fully Convolutional Network (FCN)~\cite{shelhamer2017fully} to segment the foreground object; Second, fine-tune this network based on the first frame of the video for several hundred forward-backward iterations to adapt the model to the specific video sequence. Despite the high accuracies achieved by these approaches, the fine-tuning process is arguably time consuming, which makes it prohibited for real-time applications. Some of these approaches~\cite{Cheng2017segflow}~\cite{Perazzi2017masktrack} also utilize optical flow information, which is computationally heavy for state-of-the-art algorithms\cite{Revaud2015epicflow}~\cite{Ilg2017flownet}.

In order to alleviate the computational cost of semi-supervised segmentation, we propose a novel approach to adapt the generic segmentation network to the appearance of a specific object instance in one single feed-forward pass. We propose to employ another meta neural network called \emph{modulator} to \emph{learn} to adjust the intermediate layers of the generic segmentation network given an arbitrary target object instance. Fig.~\ref{fig:overview} shows an illustration of our approach. By extracting information from the image of the annotated object and the spatial prior of the object, the modulator produces a list of parameters, which are injected into the segmentation model for layer-wise feature manipulation. Without one-shot fine-tuning, our model is able to change the behavior of the segmentation network with minimum extracted information from the target object. We name this process \emph{network modulation}.  

Our proposed model is efficient, requiring only one forward pass from the modulator to produce all parameters needed for the segmentation model to adapt to the specific object instance. Network modulation guided by the spatial prior facilitates the model to track the object even with the presence of multiple similar instances. The whole pipeline is differentiable and can be learned end-to-end using the standard stochastic gradient descent. The experiments show that our approach outperforms previous approaches without one-shot fine-tuning by a large margin, and achieves comparable performance with these approaches after one-shot fine-tuning with a 70$\times$ speed up. 

\section{Related Work}

\paragraph{Semi-supervised video segmentation.} Semi-supervised video object segmentation aims at tracking an object mask given from the first annotated frame throughout the rest of video. Many approaches have been proposed in the literature, including those propagating superpixels~\cite{Jain2014youtube}~\cite{Tsai2016objflow}, patches~\cite{jumpcut}, object proposals~\cite{objproposals}, or in bilateral space~\cite{Marki2016bilateral}, and graphical model based optimization is usually performed to consider multiple frames simultaneously. With the success of FCN on static image segmentation~\cite{Hariharan2015hypercolumns}, deep learning based methods~\cite{Perazzi2017masktrack, Caelles2017osvos, Shin2017pixel, Tokmakov2017memory, Jampani2017vpn, Cheng2017segflow} have been recently proposed for video segmentation and promising results have been achieved. To model the temporal motion information, some works heavily rely on optical flow~\cite{Tokmakov2017memory}~\cite{Cheng2017segflow}, and use CNNs to learn mask refinement of an object from current frame to the next one~\cite{Perazzi2017masktrack}, or combine the training of CNN with bilateral filtering between adjacent frames~\cite{Jampani2017vpn}. Chen~\etal~\cite{Cheng2017segflow} use a CNN to jointly estimate the optical flow and provide the learned motion representation to generate motion consistent segmentation across time. Different from these approaches, Caelles~\etal~\cite{Caelles2017osvos} combine offline and online training process on static images without using temporal information. While it saves the computation of optical flow and/or conditional random fields (CRF)~\cite{Krahenbuhl2011crf} involved in some previous methods, online fine-tuning still requires many iterations of optimization, which poses a challenge for real-world applications that need rapid inference.

\paragraph{Meta-learning for low-shot learning.} Current success of deep learning relies on the ability of learning from large-scale labeled datasets through gradient descent optimization. However, if we want our model to learn many tasks adapted to many environments, it is not affordable to learn each task for each setting from scratch. Instead, we want our deep learning system to be able to learn new tasks very fast and from very limited quantities of data. In the extreme of ``one-shot learning'', the algorithm needs to learn the new task with a single observation. One potential strategy for learning a versatile model is the notion of meta-learning, or learning to learn, which can date back to the late 80s. Recently, meta-learning has become a hot research topic with publications on neural network optimization~\cite{Chen2017LearningToLearn}, finding good network architectures, fast reinforcement learning, and few-shot image recognition \cite{Vinyals2016Matching,Ravi2016optimization,Hariharan2017ICCV,Finn2017ICML,Santoro2016ICML}. Ravi and Larochelle~\cite{Ravi2016optimization} proposed a LSTM meta-learner to learn the update rules for few shot learning. The meta optimization over a large number of tasks in \cite{Finn2017ICML} targets at learning a model that can quickly adapt to the new task with limited number of updates. Hariharan and Girschick~\cite{Hariharan2017ICCV} trained a learner that generated new samples and used new samples for training new tasks. Our approach shares the similarity with meta-learning that it learns to update the segmentation model rapidly with another meta learner, i.e. the modulator. 

\paragraph{Network manipulation}
Several previous work try to incorporate modules to manipulate the behavior of a deep neural network, either to manipulate spatial arrangement of data~\cite{Jaderberg2015spatial} or filter connections~\cite{Dai2017deformable}. Our method is also heavily motivated by conditional batch normalization~\cite{Dumoulin2017ICLR,Ghiasi2017BMVC,Huang2017ArbitraryST,Perez2017LearningVR}, where the behavior of the deep model is manipulated by batch normalization parameters conditioned on a guidance input, e.g. a style image for image stylization or a language sentence for visual question answering. 

\begin{figure*}[t]\centering
\includegraphics[width=0.85\linewidth]{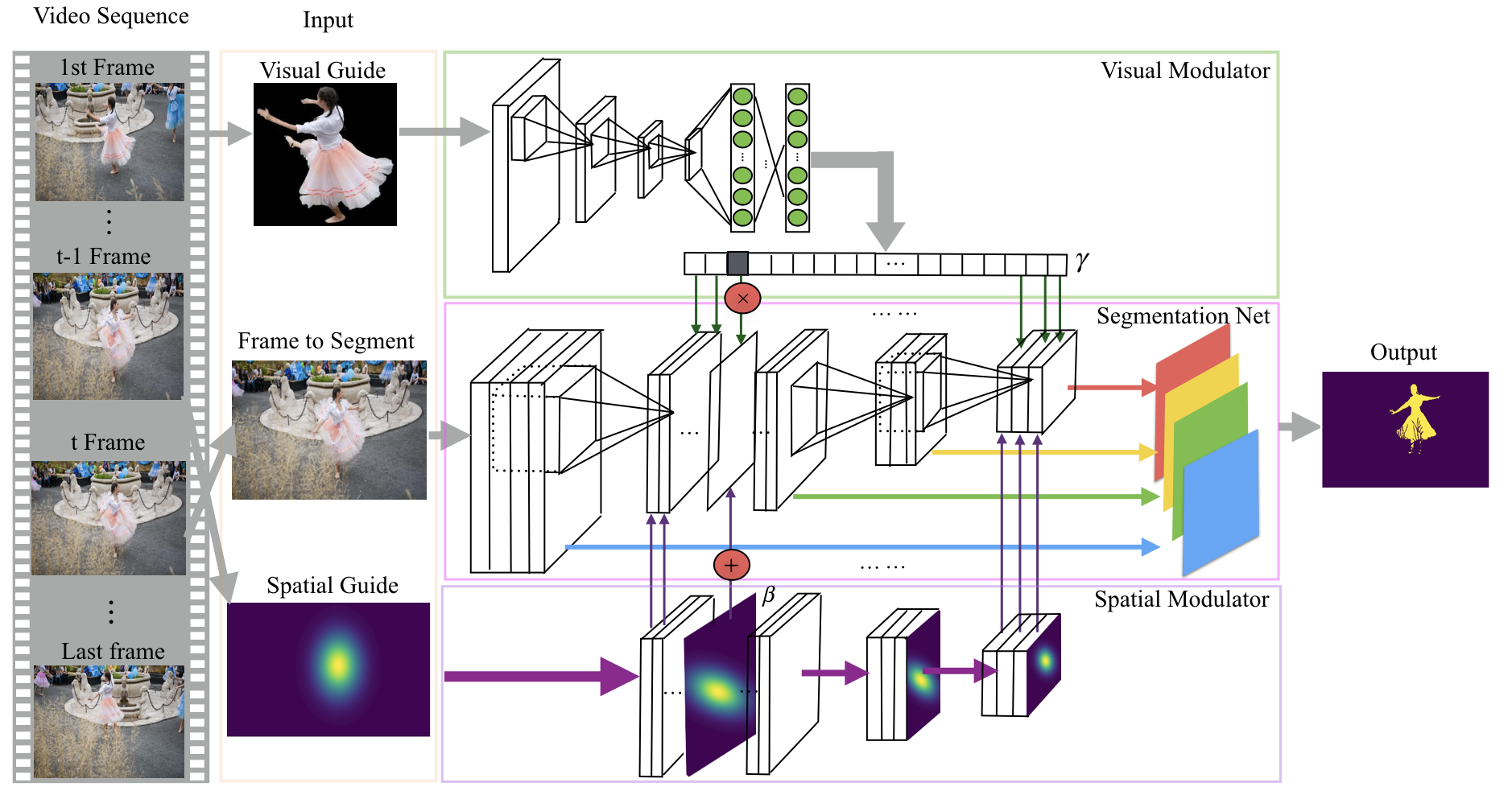}
\caption{An illustration of our model with three components: a segmentation network, a visual modulator, and a spatial modulator. The two modulators produce a set of parameters that manipulates the intermediate feature maps of the segmentation network and adapt it to segment the specific object.}
\label{fig:model}\vspace{-2pt}
\end{figure*}

\section{Video Object Segmentation with Network Modulation}

In our proposed framework, we utilize modulators to instantly adapt the segmentation network to a specific object, rather than performing hundreds of iterations of gradient descent. We can achieve similar accuracy by adjusting a limited number of parameters in the segmentation network, compared with the updating the whole network in one-shot learning approaches~\cite{Perazzi2017masktrack, Caelles2017osvos}. There are two important cues for video object segmentation: visual appearance and continuous motion in space. To use information from both visual and spatial domains, we incorporate two network modulators, namely \emph{visual modulator} and \emph{spatial modulator}, to learn to adjust intermediate layers in the main segmentation network, based on the annotated first frame and spatial location of the object, respectively.

\subsection{Conditional batch normalization}

Our approach is inspired by recent works using Conditional Batch Normalization (CBN)~\cite{Vries2017ModulatingEV,Huang2017ArbitraryST,Perez2017LearningVR}, where the scale and bias parameters of each batch-normalization layer are produced by a second controller network. These parameters are used to control the behavior of the main network for tasks such as image stylization and question answering. Mathematically, each CBN layer can be formulated as follows:
\begin{align}
\bm{y}_c = \gamma_c\bm{x}_c + \beta_c,
\end{align}
where $\bm{x}_c$ and $\bm{y}_c$ are the input and output feature maps in the $c_{th}$ channel, and $\gamma_c$ and $\beta_c$ are the scale and bias parameters produced by the controller network, respectively. The mean and variance parameters are omitted for clarity.

\subsection{Visual and spatial modulation}

The CBN layer is a special case of the more general scale-and-shift operation on feature maps. Following each convolution layer, we define a new \textit{modulation layer} with parameters generated by both visual and spatial modulators that are jointly trained. We design the two modulators such that the visual modulator produces channel-wise scale parameters to adjust the weights of different channels in the feature maps, while the spatial modulator generates element-wise bias parameters to inject spatial prior to the modulated features. Specifically, our modulation layer can be formulated as follows:
\begin{align}
\bm{y}_c = \gamma_c\bm{x}_c + \bm{\beta}_c,
\end{align}
where $\gamma_c$ and $\bm{\beta}_c$ are modulation parameters from the visual and spatial modulators, respectively. $\gamma_c$ is a scalar for channel-wise weighting, while $\bm{\beta}_c$ is a two-dimensional matrix to apply point-wise bias values. 


Fig.~\ref{fig:model} shows an illustration of the proposed approach, which consists of three networks: a fully-convolutional main segmentation network, a visual modulator network, and a spatial modulator network. The visual modulator network is a CNN that takes the annotated visual object image as input and produces a vector of scale parameters for all modulation layers, while the spatial modulator network is a very efficient network that produces bias parameters based on the spatial prior input. We will discuss the two modulators in more detail in the following sections.

\subsection{Visual modulator}

The visual modulator is used to adapt the segmentation network to focus on a specific object instance, which is the annotated object in the first frame. The annotated object is referred to as \emph{visual guide} hereafter for convenience. The visual modulator extracts semantic information such as category, color, shape, and texture, from the visual guide and generates corresponding channel-wise weights so as to re-target the segmentation network to segment the object. We use VGG16~\cite{simonyan2014very} neural network as the model for the visual modulator. We modify its last layer trained for ImageNet classification to match the number of parameters in the modulation layers for the segmentation network. 

The visual modulator implicitly learns an embedding of different types of objects. It should produce similar parameters to adjust the segmentation network for similar objects while different parameters for different objects. This is indeed true as we show in Sec.~\ref{sec:rep} that the embedding of the modulator outputs correlates with object appearance very well. One big advantage of using such a visual modulator is that we can potentially transfer the knowledge learned with a large number of object classes, e.g., ImageNet, in order to learn a good embedding.


\subsection{Spatial modulator}

Our spatial modulator takes a prior location of the object in the image as input. Since objects move continuously in a video, we set the prior to be the predicted location of the object mask in the previous frame. Specifically, we encode the location information as a heatmap with a two-dimensional Gaussian distribution on the image plane. The center and standard deviations of the Gaussian distribution are computed from the predicted mask of the previous frame. This heatmap is referred as \emph{spatial guide} hereafter for convenience. The spatial modulator downsamples the spatial guide into different scales, to match the resolution of different feature maps in the segmentation network, and then applies a scale-and-shift operation on each downsampled heatmap to generate the bias parameters of the corresponding modulation layer. Mathematically,
\begin{align}
\bm{\beta}_c = \tilde{\gamma}_c \bm{m} + \tilde{\beta}_c
\end{align}
where $\bm{m}$ is a down-sampled Gaussian heat map for the corresponding modulation layer, $\tilde{\gamma}_c$ and $\tilde{\beta}_c$ are the scale-and-shift parameters for the $c$-th channel, respectively. This is implemented with a computationally efficient $1\times 1$ convolution. In the bottom of Fig.~\ref{fig:model}, we illustrate the structure of the spatial modulator.

Our method shares some similarities with the previous work MaskTrack~\cite{Perazzi2017masktrack} in utilizing information from the previous mask. Comparing with their approach that uses the exact foreground mask of the previous frame, we only use a very coarse location prior. It may seem that our method throws away more information from the previous frame. However, we argue that the rough position and size in the previous frame possess enough information to infer the object mask with the RGB image, and it prevents the model from relying too much on the mask and as a result the error propagation, which can be catastrophic when the object has large movements in the video. As a drawback of such over-utilization of the mask, MaskTrack has to apply plenty of well-engineered data augmentation to prevent over-fitting, while we only apply simple shift and scaling as augmentation. 

\subsection{Implementation details}
 
Our FCN structure follows the one used by~\cite{Caelles2017osvos}, which is a VGG16~\cite{simonyan2014very} model with a hyper-column structure~\cite{Hariharan2015hypercolumns}.
Intuitively, we should add modulation layers after each convolution layer in the FCN. However, we found that adding modulation layers in-between the early convolution layers actually makes the model perform worse. One possible reason is that early layers extract low-level features that are very sensitive to the scale-and-shift operations introduced by the modulator. In our implementation, we add modulation operations to all convolution layers in VGG16 except the first four layers, which results in nine modulation layers.

Similar to MaskTrack~\cite{Perazzi2017masktrack}, we also utilize static images for training our model. Ideally, the visual modulator should learn a mapping from any object to modulation weights of different layers in a FCN, which requires the model to see all possible different objects. However, most video semantic segmentation datasets only contain a very limited number of categories. We tackle this challenge by using the largest public semantic segmentation dataset MS-COCO~\cite{Lin2014mscoco}, which has 80 object categories. We select objects that are larger than $3\%$ of the image size for training, resulting in a total number of $217,516$ objects. For preprocessing the input for the visual modulator, we first crop the object using the annotated mask, then set the background pixels to mean image values, and then resize the cropped image to a constant resolution of $224\times224$. The object is also augmented with up to $10\%$ random scaling and $10^{\circ}$ random rotation. For preprocessing the spatial guide as input to the spatial modulator, we first compute the mean and standard deviation of the mask, and then augment the mask with up to $20\%$ random shift and $40\%$ random scaling. For the whole image fed into the FCN, we use a random size from $320$, $400$, and $480$ with a square shape. 

The visual modulator and segmentation network are both initialized with VGG16 model pretrained on the ImageNet~\cite{Deng2009imagenet} classification task. The modulation parameters $\{\gamma_c\}$ are initialized to ones by setting the weights and biases of the last fully-connected layer of the visual modulator to zeros and ones, respectively. The weights of spatial modulator are initialized randomly. We used the same balanced cross-entropy loss as in~\cite{Caelles2017osvos}. A mini-batch size of $8$ is used. We use Adam optimizer with default momentum $0.9$ and $0.999$ for $\beta_1$ and $\beta_2$, respectively. The model is first trained for $10$ epochs with learning rate $10^{-5}$ and then trained for another $5$ epochs with learning rate $10^{-6}$.

Further, in order to model appearance variations of moving objects in videos, the model can be finetuned on video segmentation dataset such as DAVIS 2017~\cite{Pont-Tuset2017davis}. To be more robust to appearance variations, we randomly pick a foreground object from the whole video sequence as the visual guide for each frame. The spatial guide is obtained from the ground truth mask of the object in the previous frame. The same data augmentations are applied as training on MS-COCO. The model is finetuned for $20$ epochs with learning rate $10^{-6}$.


\section{Experiments}

In this section, we will introduce three parts of experiment: the comparison of our approach with previous methods, the visualization of the modulation parameters, and ablation study. Our model is trained on MS-COCO~\cite{Lin2014mscoco} 2017 dataset, and is tested on several popular video segmentation datasets, including DAVIS~\cite{Perazzi2016davis}~\cite{Pont-Tuset2017davis} and YoutubeObjects~\cite{Jain2014youtube}.

\subsection{Semi-supervised Video Segmentation}
In this section, we compare with traditional approaches including OFL~\cite{Tsai2016objflow}, BVS\cite{Marki2016bilateral}, and deep learning-based approaches including PLM~\cite{Shin2017pixel}, MaskTrack~\cite{Perazzi2017masktrack}, OSVOS~\cite{Caelles2017osvos}, VPN~\cite{Jampani2017vpn}, SFL~\cite{Cheng2017segflow}, and ConvGRU~\cite{Tokmakov2017memory}. 

\subsubsection{DAVIS 2016 \& YoutubeObjects}
First, we compare our approach with previous approaches on DAVIS 2016 and YoutubeObjects. Some approaches (MaskTrack\cite{Perazzi2017masktrack}, SFL~\cite{Cheng2017segflow} and OSVOS\cite{Caelles2017osvos}) reported results both with and without model fine-tuning on the target sequences. We include both of them and denote the variants without fine-tuning as MaskTrack-B, SFL-B, and OSVOS-B, respectively. Our model has two variants,with the first only trained on static images (Stage 1) and the second finetuned on video data (Stage 1\&2). Since there are several popular add-ons for this line of research, such as optical flow and CRF~\cite{Krahenbuhl2011crf}, which both have a lot of variants and make a fair comparison hard, we only include the performances without optical flow and CRF if possible, and mark those with add-ons in Table~\ref{tab:results}. 

\begin{table*}[t]
\centering
\caption{Performance comparison of our approach with recent approaches on DAVIS 2016 and YoutubeObjects. Performance measured in mean IU.}
\label{tab:results}
\begin{tabular}{|c|c|c|c|c|c|c|}
\hline
Method    & DAVIS 16 & YoutubeObjs & with FT & OptFlow                     & CRF                         & \begin{tabular}[c]{@{}c@{}}Speed (s)\end{tabular} \\ \hline
OFL~\cite{Tsai2016objflow}       & 68.0     & 67.5        & -       &  \cmark                           &      \cmark                       & 42.2                                                \\ \hline
BVS~\cite{Marki2016bilateral}       & 60.0     & 58.4        & -       &  \xmark                           &    \xmark                         & 0.37                                                \\ \hline
ConvGRU\cite{Tokmakov2017memory}   & 70.1     & -           & \xmark  &   \cmark                          &         \xmark                    &  20                                                   \\ \hline
VPN\cite{Jampani2017vpn}       & 70.2     & -           & \xmark  & \multicolumn{1}{c|}{\xmark} & \multicolumn{1}{c|}{\xmark} & 0.63                                                \\ \hline
MaskTrack-B~\cite{Perazzi2017masktrack} & 63.2     & 66.5        & \xmark  &  \xmark                           &       \xmark                      & 0.24                                                    \\ \hline
SFL-B~\cite{Cheng2017segflow}     & 67.4     & -        & \xmark  &   \cmark                          &        \xmark                     & 0.30                                                \\ \hline
OSVOS-B~\cite{Caelles2017osvos}     & 52.5     & 44.7        & \xmark  &   \xmark                          &        \xmark                     & 0.14                                                \\ \hline

Ours (Stage 1)      & 72.2     & 66.4        & \xmark  &  \xmark                           &       \xmark                      & 0.14 \\ \hline
Ours (Stage 1\&2)      & \textbf{74.0}     & \textbf{69.0}        & \xmark  &  \xmark                       &       \xmark                      & 0.14  \\
\hhline{|=|=|=|=|=|=|=|}
PLM~\cite{Shin2017pixel}       & 70.0     & -           & \cmark  &   \xmark                          &          \xmark                   & 0.50                                                   \\ \hline
MaskTrack~\cite{Perazzi2017masktrack} & 69.8     & 71.7        & \cmark  &   \xmark                          &      \xmark                       & 12                                                  \\ \hline
SFL~\cite{Cheng2017segflow}     & 74.8     & -        & \cmark  &   \cmark                          &        \xmark                     & 7.9                                                \\ \hline
OSVOS~\cite{Caelles2017osvos}     & \textbf{79.8}     & \textbf{74.1}        & \cmark  & \xmark                            &       \xmark                      & 10                                                  \\ \hline
\end{tabular}
\end{table*}

In Table~\ref{tab:results}, by comparing our method with OFL~\cite{Tsai2016objflow}, an expensive graphical model based approach, we achieve better accuracy on both DAVIS 2016 and YoutubeObjects. Comparing with deep learning approaches without model fine-tuning, and therefore, similar speed as ours, our method achieves the best accuracy on both DAVIS 2016 and YoutubeObjects. Comparing with the four approaches using model fine-tuning on target videos (PLM, MaskTrack, SFL, and OSVOS), our approach achieves better performance than PLM and MaskTrack, and is on-par with SFL. OSVOS achieves higher accuracy but it also utilizes a boundary snapping approach which contributes $2.4\%$ in mean IU. Our method is $70\times$ faster than MaskTrack and OSVOS, $50\times$ faster than SFL. We measure the running time of MaskTrack-B, OSVOS-B, and our method on a NVIDIA Quadro M6000 GPU using Tensorflow~\cite{tensorflow2015}. Speed of other methods are derived from the corresponding papers~\footnote{Speed of ConvGRU is estimated with the expensive optical flow they use, speed of PLM is derived through communication with the authors.}.

In our method, the adaptation of the segmentation model by the modulators is done with one forward pass for visual modulator, so it is much more efficient than the approaches with model fine-tuning on target videos. The visual modulator only needs to be computed once for the whole video, while the spatial modulator needs to be computed for every frame but the overhead is negligible, i.e., the average speed of our model on a video sequence is about the same as FCN itself. Our method is the second fastest of all compared methods, with only MaskTrack-B and OSVOS-B achieving similar speed but with much worse accuracies.

\subsubsection{DAVIS 2017}
\begin{figure*}[t]\centering
\includegraphics[width=0.9\linewidth]{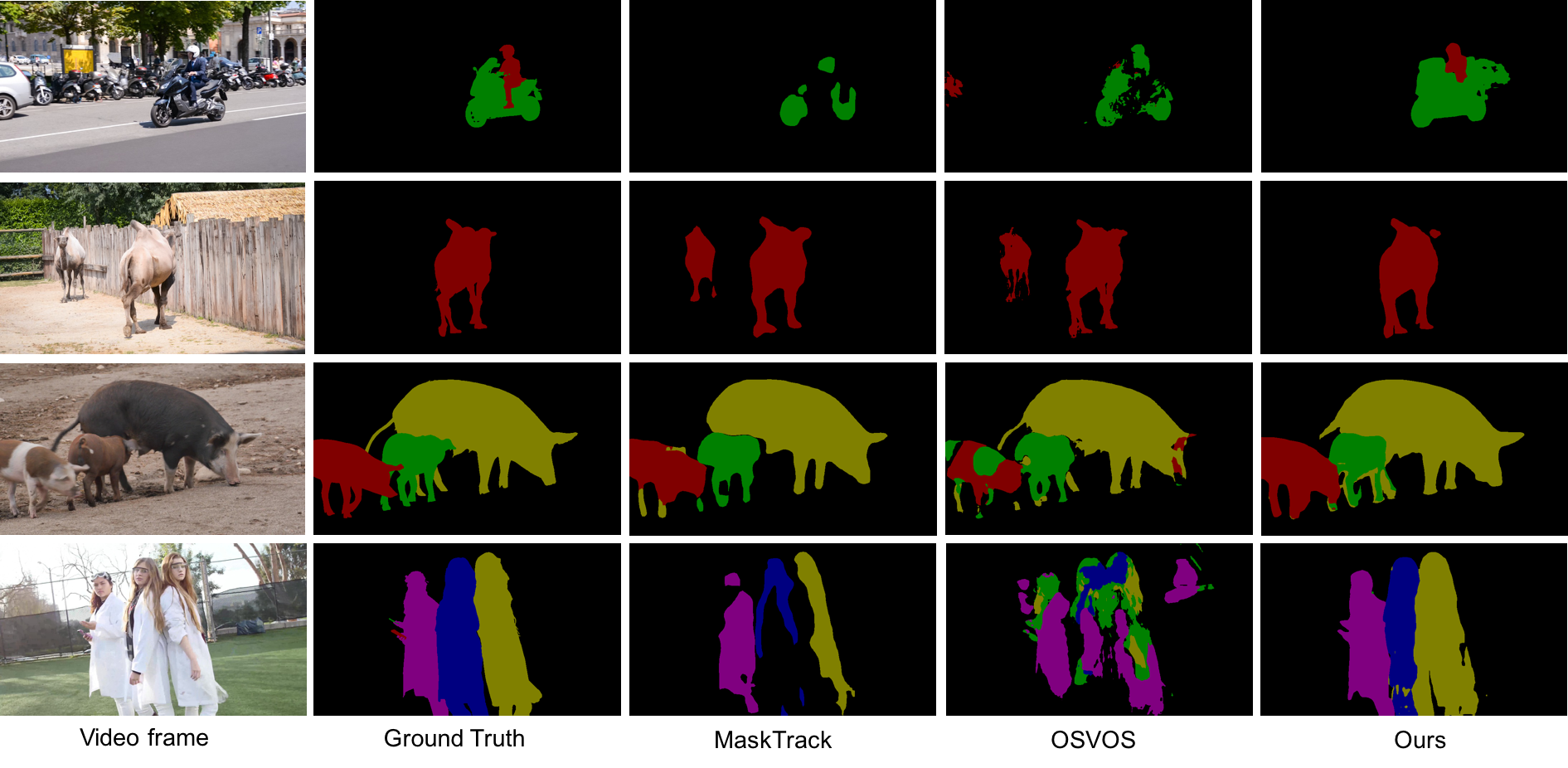}
\caption{Some qualitative results of our approach compared with two recent state-of-the-art approaches on DAVIS 2017.}
\label{fig:compare}\vspace{-2pt}
\end{figure*}

To further investigate the capability of our model, we conduct more experiments on DAVIS 2017~\cite{Pont-Tuset2017davis}, which is the largest video segmentation dataset to date. DAVIS 2017 is more challenging than DAVIS 2016 and YoutubeObjects in that it has multiple objects for each video sequences and some of the objects are very similar. We compare our method with two most related approaches, MaskTrack~\cite{Perazzi2017masktrack} and OSVOS~\cite{Caelles2017osvos}. For fair comparison, we only use their single network and adds-on free versions. We directly use open source code of OSVOS and adapt MaskTrack model to Tensorflow~\cite{tensorflow2015}. For each video sequence, OSVOS and MaskTrack are finetuned with $1000$ iterations. To show that network modulation is capable of adapting different model structures to specific object instances, we also experiment with modified OSVOS and MaskTrack models by adding a visual modulator to each of them, which are named OSVOS-M and MaskTrack-M respectively. For these two models, we only update the weights of the visual modulators and keep the weights of the segmentation model fixed in training. 

\begin{figure}[t]\centering
\includegraphics[width=1\linewidth]{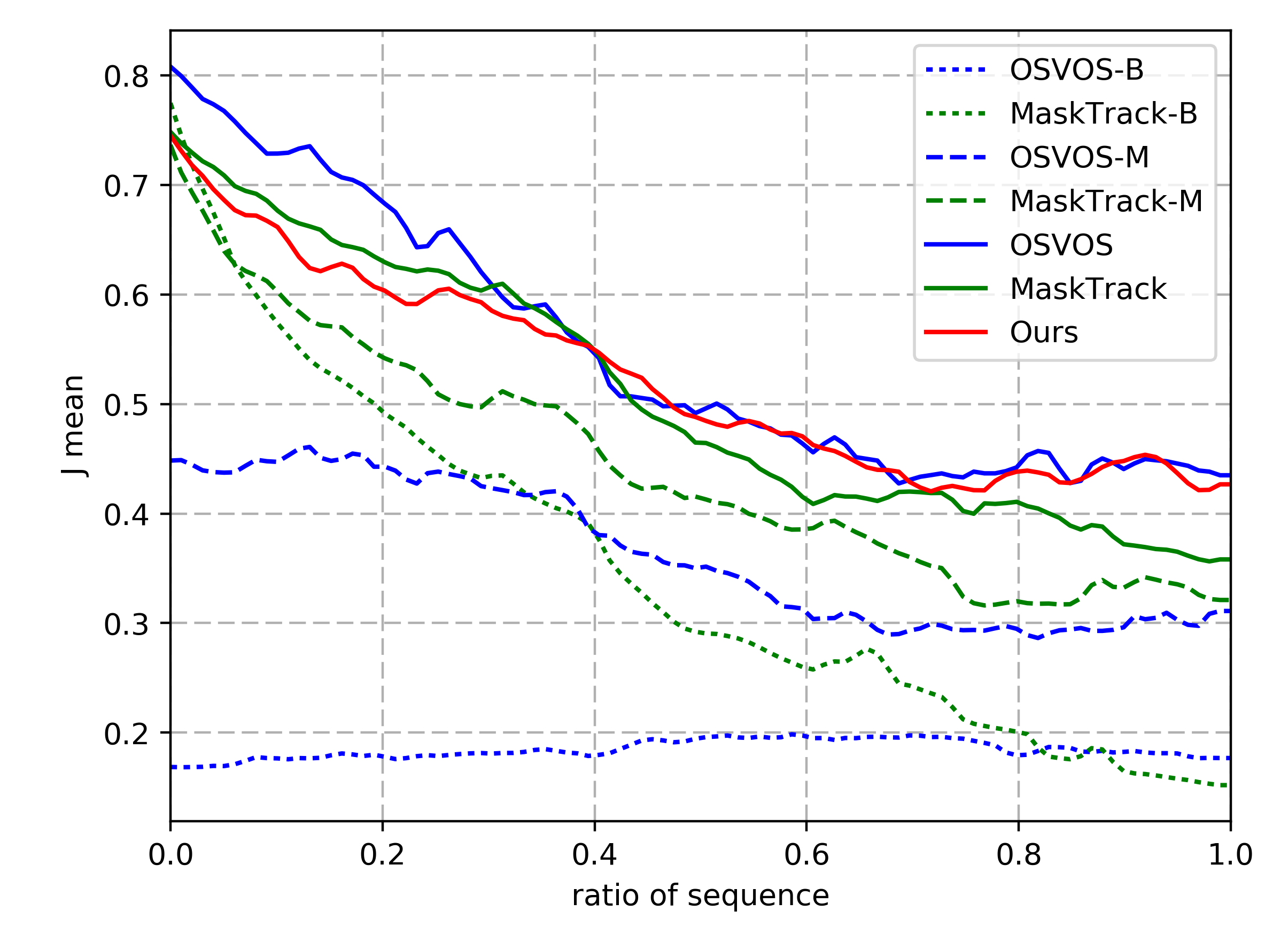}
\caption{The $\mathcal{J}$ mean performance of different methods over time on DAVIS 2017. Best viewed in color.}
\label{fig:over_time}\vspace{-2pt}
\end{figure}

Table~\ref{tab:res_2017} shows the results of different approaches on DAVIS 2017. We utilize the official evaluation metrics of DAVIS dataset: mean, recall, and decay of region similarity $\mathcal{J}$ and contour accuracy $\mathcal{F}$, respectively. Note $\mathcal{J}$ mean is equivalent to mean IU we used above. Again, our model outperforms OSVOS-B and MaskTrack-B with a large margin, while obtaining comparable performance with the two methods with model fine-tuning. OSVOS-M and MaskTrack-M are both better than their baseline implementations with a $18\%$ and $9.3\%$ gain in $\mathcal{J}$ mean, respectively. Since the weights of the segmentation model are fixed, the accuracy gain comes solely from the modulator, which proves that the visual modulator is capable of improving different model structures by manipulating the scales of the intermediate feature maps. Noticeably, our method obtains much lower decay rate for both region similarity and contour accuracy compared to OSVOS and MaskTrack. The accuracy changes of the different methods over time are illustrated in Fig.~\ref{fig:over_time}. In the beginning of the video, our method lags behind OSVOS and MaskTrack. However, when it proceeds to around $40\%$ of the video, our method is on par with OSVOS and outperforms MaskTrack towards the end of the video. With one-shot fine-tuning, OSVOS and MaskTrack fit to the first frame very well. They are able to obtain high accuracy in the beginning of the video since these frames are all similar to the first one. But as time goes on and the object turns into different poses and appearances, it gets harder for the fine-tuned model to generalize to new object appearances. Our model is more robust to the appearance changes since it learns a feature embedding (see Section~\ref{sec:rep}) for the annotated object which is more tolerant to pose and appearance changes compared to one-shot fine-tuning.
                                                                                                                                                                                                       
Some qualitative results of our methods compared with the two previous approaches are shown in Fig.~\ref{fig:compare}. Compared with MaskTrack, our method generally obtains more accurate boundaries, partially due to that the coarse spatial prior forces the model to explore more cues on the image rather than the mask in the previous frame. Compared with OSVOS, our method shows better results when there are multiple similar objects in the image, thanks to the tracking capability provided by the spatial modulator. On the other hand, our method is also shown to work well on unseen object categories in training data. In Fig.~\ref{fig:compare}, the camel and the pigs are unseen object categories in MS-COCO dataset.

\begin{table*}[t]
\centering
\caption{Comparisons of our approach and two state-of-the-art algorithm on DAVIS 2017 validation set.}
\label{tab:res_2017}
\begin{tabular}{|l|l|l|l|l|l|l|l|}
\hline
Method      & with FT & $\mathcal{J}$ mean$\uparrow$ & $\mathcal{J}$ recall$\uparrow$ & $\mathcal{J}$ decay$\downarrow$ & $\mathcal{F}$ mean$\uparrow$ & $\mathcal{F}$  recall$\uparrow$ & $\mathcal{F}$  decay$\downarrow$ \\ \hline
OSVOS-B~\cite{Caelles2017osvos}       & \xmark  & 18.5   & 15.9     & -0.8    & 30.0   & 20.0     & 0.1     \\ \hline
MaskTrack-B~\cite{Perazzi2017masktrack}   & \xmark  & 35.3   & 37.8     & 39.3    & 36.4   & 36.0     & 42.0    \\ \hline
OSVOS-M     & \xmark  & 36.4    & 34.8    &14.8    & 39.5     & 35.3     & 9.1    \\ \hline
MaskTrack-M & \xmark  & 44.6   & 48.7      & 27.1     & 47.6   & 49.3     & 27.9     \\ \hline
OSVOS~\cite{Caelles2017osvos}        & \cmark  & 55.1   & 60.2     & 28.2    & 62.1   & 71.3     & 29.3    \\ \hline
MaskTrack~\cite{Perazzi2017masktrack}   & \cmark  & 51.2   & 59.7     & 28.3    & 57.3   & 65.5     & 29.1    \\ \hline
Ours        & \xmark  & 52.5   & 60.9     & 21.5    & 57.1   & 66.1     & 24.3    \\ \hline
\end{tabular}
\end{table*}

\begin{figure}[t]\centering
\includegraphics[width=1.0\linewidth]{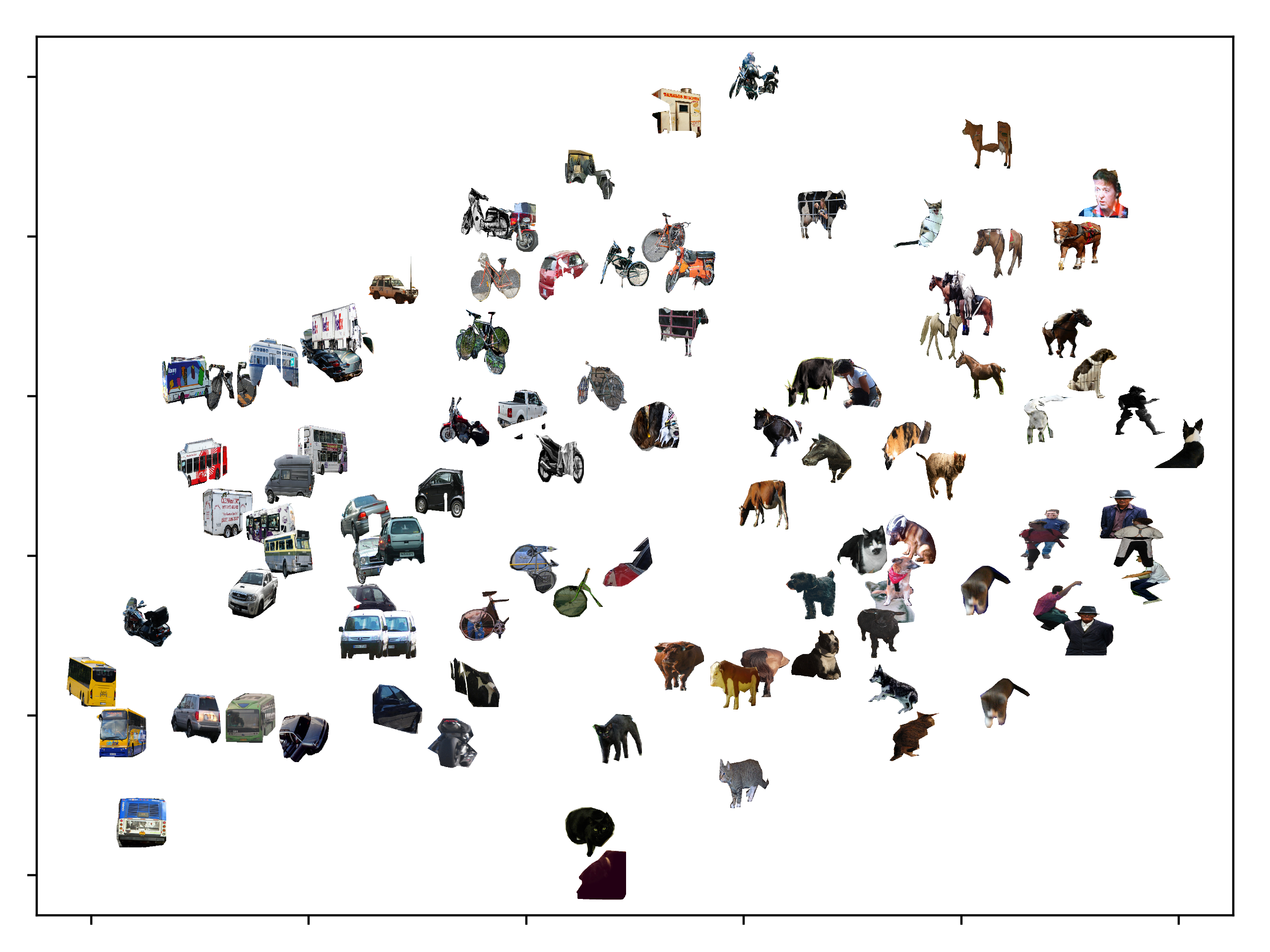}
\caption{Visualization of learned modulation parameters for 100 objects from 10 categories: bicycle, motorcycle, car, bus, truck, dog, cat, horse, cow, person. Zoom in to see details.}
\label{fig:visual}\vspace{-2pt}
\end{figure}

\begin{figure}[t]\centering
\includegraphics[width=1.0\linewidth]{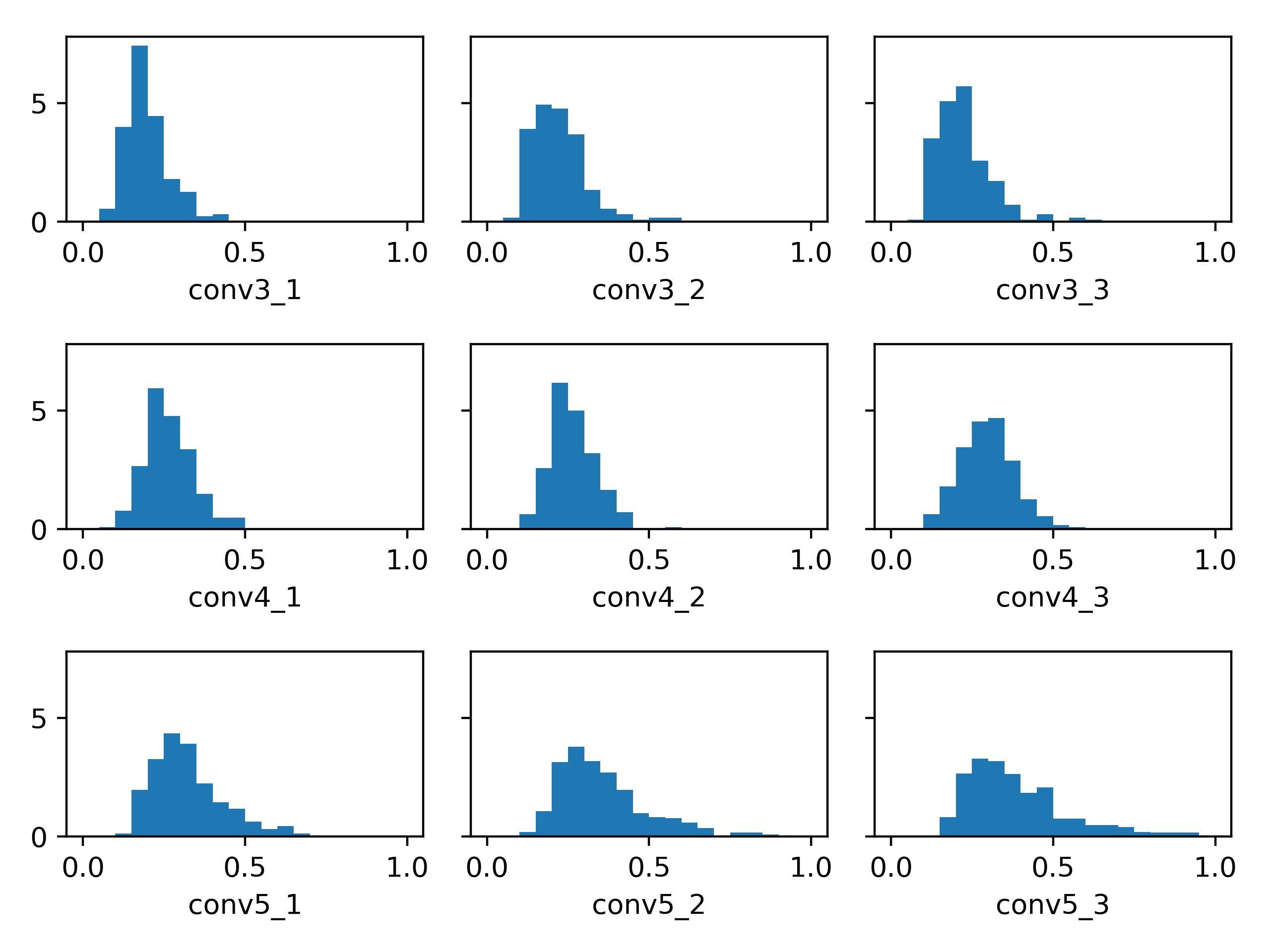}
\caption{Histograms of standard deviations of $\gamma_c$ from the visual modulator in different modulation layers. The annotated names are the corresponding convolution layers in VGG16.}
\label{fig:std_vis}\vspace{-2pt}
\end{figure}

\begin{figure}[t]\centering
\includegraphics[width=1.0\linewidth]{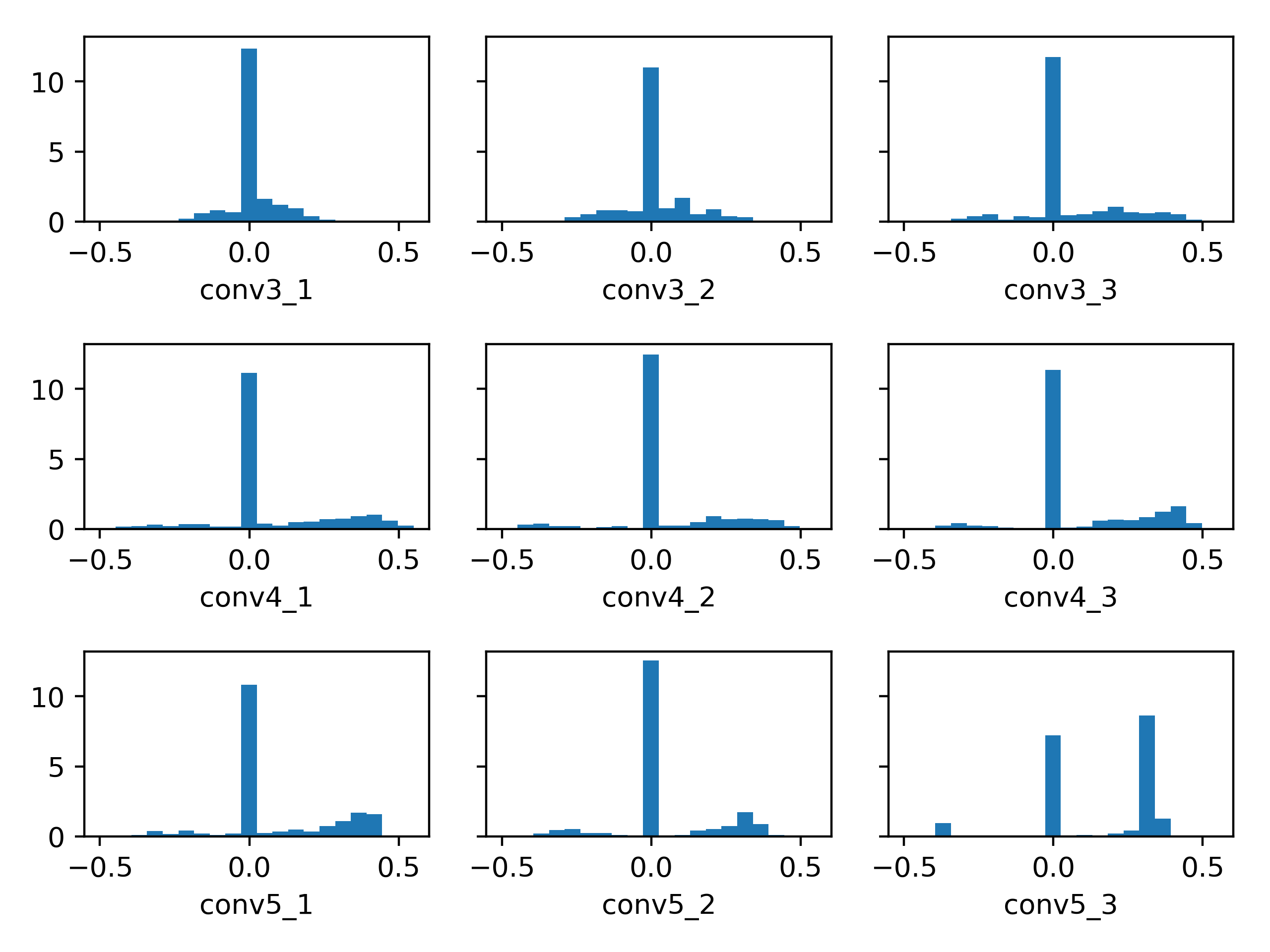}
\caption{Histograms of magnitude of $\tilde{\gamma}_c$ from the spatial modulator in different modulation layers. The annotated names are the corresponding convolution layers in VGG16.}
\label{fig:mag_sp}\vspace{-2pt}
\end{figure}

\subsection{Visualization of the modulation parameters}\label{sec:rep}
Our model implicitly learns an embedding with the modulation parameters from the visual modulator for the annotated objects. Intuitively, similar objects should have similar modulation parameters, while different objects should have dramatically different modulation parameters. To visualize this embedding, we extract modulation parameters from $100$ object instances in $10$ object classes in MS-COCO, and visualize the parameters in a two-dimensional embedding space using multi-dimensional scaling in Fig.~\ref{fig:visual}. We can see that objects in the same category are mostly clustered together, and similar categories are closer to each other than dissimilar categories. For example, cats and dogs, cars and buses are mixed up due to their similar appearance, while bicycles and dogs, buses and horses are far from each other due to the big visual difference. Mammal classes (cats, dogs, cows, horses, human) are generally clustered together, and man-made objects (cars, buses, bicycles, motorcycles, trucks) are clustered together. 

We also investigate the magnitude of the modulation parameters in different layers. The modulation parameters $\{\gamma_c\}$ changes according to the visual guide. Therefore, we compute the standard deviations of modulation parameters $\{\gamma_c\}$ in each modulation layer for images in MS-COCO validation set and illustrate them in Fig.~\ref{fig:std_vis}. An interesting observation is that towards deeper level of the network, the variations of modulation parameters get larger. This shows that the manipulation of feature maps is more dramatic in the last few layers than in early layers of the network. The last few layers of a deep neural network usually learn high-level semantic meanings~\cite{Zeiler2014visualizing}, which could be used to adjust the segmentation model to a specific object more effectively.

We also look into the spatial modulator by extracting the scale parameters $\{\tilde{\gamma}_c\}$ in each layer of the spatial modulator and visualize them in Fig.~\ref{fig:mag_sp}. The magnitudes of $\{\tilde{\gamma}_c\}$ are the relative scales of the spatial guide added to the feature maps in the FCN. The scale of $\{\tilde{\gamma}_c\}$ is proportional to the impact of spatial prior on the intermediate feature maps. Interestingly, we observe sparsity in the values of $\{\tilde{\gamma}_c\}$. Except the last convolution layer \texttt{conv5\_3}, around $60\%$ of the parameters have zero values, which means only $40\%$ of the feature maps are affected by the spatial prior in these layers. In the layer \texttt{conv5\_3}, around $70\%$ of the feature maps interact with the spatial guide and most of them are added with a similar scale (note the peak around $0.4$) of the spatial guide. This shows that the spatial prior is fused into the feature maps gradually, rather than being effective at the beginning of the network. After all feature extractions are done, the spatial modulator makes a large adjustment to the feature maps, which provides a strong prior of the location of the target object.

\begin{table}[]
\centering
\caption{Ablation study of our method on DAVIS 2017.}
\label{tab:ablation}
\begin{tabular}{llll}
\hline
                       & Variants                      & mIU  & $\Delta$ mIU \\ \hline
\multirow{2}{*}{Add-on}     & Ours + Online finetuning             & \textbf{60.8} & +8.3        \\ \cline{2-4}
                  & Ours + CRF                    &  54.4    &  +1.9            \\ \hline
                       & Ours                          & 52.5 &              \\ \hline
\multirow{2}{*}{Model} & no visual modulator           & 33.0 &   -19.5           \\ \cline{2-4} 
                       & no spatial modulator          & 40.1 &   -12.4           \\ \hline
\multirow{3}{*}{Data}  & - random crop  & 50.6 &      -1.9        \\ 
                       & - visual guide augmentation &  49.5    &  -1.1 \\
					& - spatial guide augmentation & 35.6  &  -13.9                     \\ \hline
\end{tabular}
\end{table}

\subsection{Ablation Study}
We study the impact of different ingredients in our method. We conduct experiments on DAVIS 2017 and measure the performance using mean IU. For variants of model structures, we experiment with only using spatial or visual modulator. For data augmentation methods, we experiment with no random crop augmentation for the FCN input, and no affine transformation for the visual guide and the spatial guide. We experiment with CRF as a post-processing step. To investigate the effect of one-shot fine-tuning on our model, we also experiment with standard one-shot fine-tuning using a small number of iterations. Results are shown in Table~\ref{tab:ablation}.

By adding a CRF post-processing, our method achieves mIU (mean IU) of $54.4$. By one-shot fine-tuning with only $100$ iterations for each sequence, our method achieves mIU of $60.8$, which is $5.7$ better than OSVOS with $1000$ iterations. With fine-tuning, our method is still relatively efficient with average running time around $1$ s/frame. Without visual modulator, our model deteriorates to $33.0$, while without spatial modulator, our model obtains mIU of $40.1$, which shows that the visual guide is more important than the spatial guide. For data augmentation, without random crop, the accuracy drops by $1.9$. Without affine data augmentation on the visual guide, the accuracy further decreases by $1.1$. Without augmentation on the spatial guide, our model only obtains mIU of $35.6$, which is a dramatic drop from $49.5$. The results indicates that the spatial guide augmentation is the most significant on the performance. Without perturbation, the model might rely on the location of the spatial prior too much that it cannot deal with moving objects in real video sequences.

\section{Conclusions}
In this work, we propose a novel framework to process one-shot video segmentation efficiently. To alleviate the slow speed of one-shot fine-tuning developed by previous FCN-based methods, we propose to use a network modulation approach mimicking the fine-tuning process with one forward pass of the modulator network. We show in experiments that by injecting a limited number of parameters computed by the modulators, the segmentation model can be re-purposed to segment an arbitrary object. The proposed network modulation method is a general learning method for few-shot learning problems, which could be applied to other tasks such as visual tracking and image stylization. Our approach falls into the general category of meta-learning, and it would also be interesting to investigate other meta-learning approaches for video segmentation. Another piece of future work would be to learn a recurrent representation of the modulation parameters to manipulate the FCN based on temporal information. 

{\small
\bibliographystyle{ieee}
\bibliography{egbib}

\begin{thebibliography}{10}\itemsep=-1pt

\bibitem{tensorflow2015}
M.~Abadi, A.~Agarwal, P.~Barham, E.~Brevdo, Z.~Chen, C.~Citro, G.~S. Corrado,
  A.~Davis, J.~Dean, M.~Devin, S.~Ghemawat, I.~Goodfellow, A.~Harp, G.~Irving,
  M.~Isard, Y.~Jia, R.~Jozefowicz, L.~Kaiser, M.~Kudlur, J.~Levenberg,
  D.~Man\'{e}, R.~Monga, S.~Moore, D.~Murray, C.~Olah, M.~Schuster, J.~Shlens,
  B.~Steiner, I.~Sutskever, K.~Talwar, P.~Tucker, V.~Vanhoucke, V.~Vasudevan,
  F.~Vi\'{e}gas, O.~Vinyals, P.~Warden, M.~Wattenberg, M.~Wicke, Y.~Yu, and
  X.~Zheng.
\newblock {TensorFlow}: Large-scale machine learning on heterogeneous systems,
  2015.
\newblock Software available from tensorflow.org.

\bibitem{Caelles2017osvos}
S.~Caelles, K.-K. Maninis, J.~Pont-Tuset, L.~Leal-Taix\'e, D.~Cremers, and
  L.~{Van Gool}.
\newblock One-shot video object segmentation.
\newblock In {\em CVPR}, 2017.

\bibitem{Chen2017LearningToLearn}
Y.~Chen, M.~W. Hoffman, S.~G. Colmenarejo, M.~Denil, T.~P. Lillicrap,
  M.~Botvinick, and N.~de~Freitas.
\newblock Learning to learn without gradient descent by gradient descent.
\newblock In {\em ICML}, 2016.

\bibitem{Cheng2017segflow}
J.~Cheng, Y.-H. Tsai, S.~Wang, and M.-H. Yang.
\newblock Segflow: Joint learning for video object segmentation and optical
  flow.
\newblock In {\em IEEE International Conference on Computer Vision (ICCV)},
  2017.

\bibitem{Dai2017deformable}
J.~Dai, H.~Qi, Y.~Xiong, Y.~Li, G.~Zhang, H.~Hu, and Y.~Wei.
\newblock Deformable convolutional networks.
\newblock {\em ICCV}, 2017.

\bibitem{Vries2017ModulatingEV}
H.~de~Vries, F.~Strub, J.~Mary, H.~Larochelle, O.~Pietquin, and A.~C.
  Courville.
\newblock Modulating early visual processing by language.
\newblock {\em CoRR}, abs/1707.00683, 2017.

\bibitem{Deng2009imagenet}
J.~Deng, W.~Dong, R.~Socher, L.-J. Li, K.~Li, and L.~Fei-Fei.
\newblock Imagenet: A large-scale hierarchical image database.
\newblock In {\em CVPR}, 2009.

\bibitem{Dumoulin2017ICLR}
V.~Dumoulin, J.~Shlens, and M.~Kudlur.
\newblock A learned representation for artistic style, 2017.

\bibitem{jumpcut}
Q.~Fan, F.~Zhong, D.~Lischinski, D.~Cohen-Or, and B.~Chen.
\newblock Jumpcut:non-successive mask transfer and interpolation for video
  cutout.
\newblock In {\em ACM Trans. Graph., 34(6)}, 2015.

\bibitem{Finn2017ICML}
C.~Finn, P.~Abbeel, and S.~Levine.
\newblock Model-agnostic meta-learning for fast adaptation of deep networks.
\newblock In {\em ICML}, 2017.

\bibitem{Ghiasi2017BMVC}
G.~Ghiasi, H.~Lee, M.~Kudlur, V.~Dumoulin, and J.~Shlens.
\newblock exploring the structure of a real-time, arbitrary neural artistic
  stylization network.
\newblock In {\em BMVC}, 2017.

\bibitem{Hariharan2015hypercolumns}
B.~Hariharan, P.~Arbel{\'a}ez, R.~Girshick, and J.~Malik.
\newblock Hypercolumns for object segmentation and fine-grained localization.
\newblock In {\em CVPR}, 2015.

\bibitem{Hariharan2017ICCV}
B.~Hariharan and R.~Girshick.
\newblock Low-shot visual recognition by shrinking and hallucinating features.
\newblock In {\em ICCV}, 2017.

\bibitem{Huang2017ArbitraryST}
X.~Huang and S.~J. Belongie.
\newblock Arbitrary style transfer in real-time with adaptive instance
  normalization.
\newblock {\em CoRR}, abs/1703.06868, 2017.

\bibitem{Ilg2017flownet}
E.~Ilg, N.~Mayer, T.~Saikia, M.~Keuper, A.~Dosovitskiy, and T.~Brox.
\newblock Flownet 2.0: Evolution of optical flow estimation with deep networks.
\newblock {\em CVPR}, 2017.

\bibitem{Jaderberg2015spatial}
M.~Jaderberg, K.~Simonyan, A.~Zisserman, et~al.
\newblock Spatial transformer networks.
\newblock In {\em NIPS}, pages 2017--2025, 2015.

\bibitem{Jain2014youtube}
S.~D. Jain and K.~Grauman.
\newblock Supervoxel-consistent foreground propagation in video.
\newblock In {\em ECCV}, 2014.

\bibitem{Jampani2017vpn}
V.~Jampani, R.~Gadde, and P.~V. Gehler.
\newblock Video propagation networks.
\newblock In {\em CVPR}, 2017.

\bibitem{Krahenbuhl2011crf}
P.~Kr{\"a}henb{\"u}hl and V.~Koltun.
\newblock Efficient inference in fully connected crfs with gaussian edge
  potentials.
\newblock In {\em NIPS}, pages 109--117, 2011.

\bibitem{Li2013video}
F.~Li, T.~Kim, A.~Humayun, D.~Tsai, and J.~M. Rehg.
\newblock Video segmentation by tracking many figure-ground segments.
\newblock In {\em ICCV}, 2013.

\bibitem{Lin2014mscoco}
T.-Y. Lin, M.~Maire, S.~Belongie, J.~Hays, P.~Perona, D.~Ramanan,
  P.~Doll{\'a}r, and C.~L. Zitnick.
\newblock Microsoft coco: Common objects in context.
\newblock In {\em ECCV}, 2014.

\bibitem{Marki2016bilateral}
N.~M{\"a}rki, F.~Perazzi, O.~Wang, and A.~Sorkine-Hornung.
\newblock Bilateral space video segmentation.
\newblock In {\em CVPR}, 2016.

\bibitem{Perazzi2017masktrack}
F.~Perazzi, A.~Khoreva, R.~Benenson, B.~Schiele, and A.Sorkine-Hornung.
\newblock Learning video object segmentation from static images.
\newblock In {\em CVPR}, 2017.

\bibitem{Perazzi2016davis}
F.~Perazzi, J.~Pont-Tuset, B.~McWilliams, L.~{Van Gool}, M.~Gross, and
  A.~Sorkine-Hornung.
\newblock A benchmark dataset and evaluation methodology for video object
  segmentation.
\newblock In {\em CVPR}, 2016.

\bibitem{objproposals}
F.~Perazzi, O.~Wang, M.~Gross, and A.~Sorkine-Hornung.
\newblock Fully connected object proposals for video segmentation.
\newblock In {\em ICCV}, 2015.

\bibitem{Perez2017LearningVR}
E.~Perez, H.~de~Vries, F.~Strub, V.~Dumoulin, and A.~C. Courville.
\newblock Learning visual reasoning without strong priors.
\newblock {\em CoRR}, abs/1707.03017, 2017.

\bibitem{Pont-Tuset2017davis}
J.~Pont-Tuset, F.~Perazzi, S.~Caelles, P.~Arbel\'aez, A.~Sorkine-Hornung, and
  L.~{Van Gool}.
\newblock The 2017 davis challenge on video object segmentation.
\newblock {\em arXiv:1704.00675}, 2017.

\bibitem{Ravi2016optimization}
S.~Ravi and H.~Larochelle.
\newblock Optimization as a model for few-shot learning.
\newblock {\em ICLR}, 2017.

\bibitem{Revaud2015epicflow}
J.~Revaud, P.~Weinzaepfel, Z.~Harchaoui, and C.~Schmid.
\newblock Epicflow: Edge-preserving interpolation of correspondences for
  optical flow.
\newblock In {\em CVPR}, 2015.

\bibitem{Santoro2016ICML}
A.~Santoro, S.~Bartunov, M.~Botvinick, D.~Wierstra, and T.~Lillicrap.
\newblock Meta-learning with memory-augmented neural networks.
\newblock In {\em ICML}, 2016.

\bibitem{shelhamer2017fully}
E.~Shelhamer, J.~Long, and T.~Darrell.
\newblock Fully convolutional networks for semantic segmentation.
\newblock {\em IEEE transactions on pattern analysis and machine intelligence},
  39(4):640--651, 2017.

\bibitem{Shin2017pixel}
J.~Shin~Yoon, F.~Rameau, J.~Kim, S.~Lee, S.~Shin, and I.~So~Kweon.
\newblock Pixel-level matching for video object segmentation using
  convolutional neural networks.
\newblock In {\em CVPR}, 2017.

\bibitem{simonyan2014very}
K.~Simonyan and A.~Zisserman.
\newblock Very deep convolutional networks for large-scale image recognition.
\newblock {\em arXiv preprint arXiv:1409.1556}, 2014.

\bibitem{Tokmakov2017memory}
P.~Tokmakov, K.~Alahari, and C.~Schmid.
\newblock Learning video object segmentation with visual memory.
\newblock In {\em ICCV}, 2017.

\bibitem{Tsai2016objflow}
Y.-H. Tsai, M.-H. Yang, and M.~J. Black.
\newblock Video segmentation via object flow.
\newblock In {\em CVPR}, 2016.

\bibitem{Vinyals2016Matching}
O.~Vinyals, C.~Blundell, T.~Lillicrap, K.~Kavukcuoglu, and D.~Wierstra.
\newblock Matching networks for one shot learning.
\newblock In {\em NIPS}, 2016.

\bibitem{Zeiler2014visualizing}
M.~D. Zeiler and R.~Fergus.
\newblock Visualizing and understanding convolutional networks.
\newblock In {\em ECCV}, 2014.

\end{thebibliography}
}

\end{document}